\documentclass{article}

\usepackage[preprint]{neurips_2020}
\usepackage{longtable}
\usepackage[utf8]{inputenc} % allow utf-8 input
\usepackage[T1]{fontenc}    % use 8-bit T1 fonts
\usepackage{hyperref}       % hyperlinks
\usepackage{url}            % simple URL typesetting
\usepackage{booktabs}       % professional-quality tables
\usepackage{amsfonts}       % blackboard math symbols
\usepackage{nicefrac}       % compact symbols for 1/2, etc.
\usepackage{microtype}      % microtypography
\usepackage{xspace}

\newcommand{\margenews}{\textsc{MARGE-NEWS}\xspace}
\newcommand{\marge}{\textsc{MARGE}\xspace}
\usepackage{amsmath}
\usepackage{graphicx}
\usepackage{color}
\newcommand{\norm}[1]{\left\lVert #1 \right\rVert}
\usepackage{subcaption}
\usepackage[utf8]{inputenc}
\usepackage{comment}

\usepackage{endnotes}

\title{Pre-training via Paraphrasing}

\author{%
  Mike Lewis \And Marjan Ghazvininejad \And Gargi Ghosh \AND Armen Aghajanyan \And Sida Wang \\ \\
  Facebook AI \\
  \texttt{mikelewis@fb.com} \And Luke Zettlemoyer  \\
}

\begin{document}

\maketitle

% Key Points:
%-
% First pre-training that isn't MLM to work really well
% Paraphrasiung is a crucial skill for many applications
% Multi-lingual multi-document
% Clean: retrieval model and language model optimized from a single training objective and random initialisation
% Pre-training closely related to downstream tasks, allowing zero-shot performance (connect to probing?)
% Good performance on discriminative and generative tasks in many languages - most general pretrained model yet?
% New capabilites? e.g. summarise document collection in a different language

\begin{figure}[h]
    \centering
    \includegraphics[width=\textwidth]{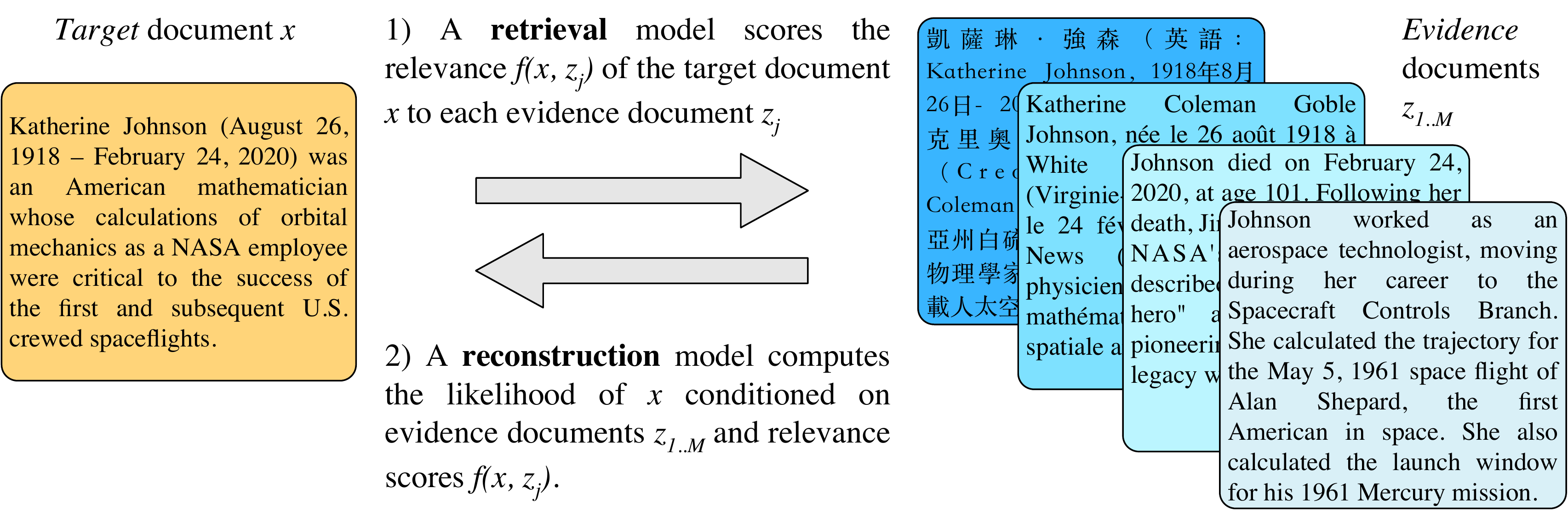}
    %\caption{Overview of \marge: documents are predicted conditioned on a set of related documents. A retrieval model is used to both find batches of evidence documents, and to bias the reconstruction model to pay attention to the most relevant evidence documents. Optimizing the reconstruction loss teaches the model both paraphrasing and retrieval skills.}
    \label{fig:my_label}
    \vspace{-10pt}
    \caption{Pre-training via Paraphrasing: a retrieval model maps a document to a set of related documents, which a reconstruction model paraphrases to maximize the likelihood of the original. %$^\dagger$ 
    Example text adapted from {\small \url{https://{en,es,de,it,fr,zh}.wikipedia.org/wiki/Katherine_Johnson}}}
    \vspace{5pt}
\end{figure}
%\endnotetext{TODO}
%\footnotetext{\url{{en,fr,zh}.wikipedia.org/wiki/barack_obama} and \url{en.wikipedia.org/wiki/Presidency_of_Barack_Obama}}

\begin{abstract}

We introduce \marge, a pre-trained sequence-to-sequence model learned with an unsupervised multi-lingual multi-document paraphrasing objective. \marge provides an alternative to the dominant masked language modeling paradigm, where we self-supervise the \emph{reconstruction} of target text by \emph{retrieving} a set of related texts (in many languages) and conditioning on them to maximize the likelihood of generating the original. We show it is possible to jointly learn to do retrieval and reconstruction, given only a random initialization. The objective noisily captures aspects of paraphrase, translation, multi-document summarization, and information retrieval, allowing for strong zero-shot performance on several tasks. For example, with no additional task-specific training we achieve BLEU scores of up to 35.8 for document translation. We further show that fine-tuning gives strong performance on a range of discriminative and generative tasks in many languages, making \marge the most generally applicable pre-training method to date.
%providing the first competitive alternative to masked language models.

\end{abstract}

\section{Introduction}
Variations on masked language models (MLMs) \citep{bert,roberta,xlnet,xlmr,bart,t5,electra} provide highly effective self supervision for pre-training by removing and then reconstructing parts of an input text. In this paper, we present the first viable pretraining alternative to MLMs; self supervision is instead provided by learning to paraphrase collections of related documents in many languages. 

More specifically, we introduce \marge, a \textbf{M}ultilingual \textbf{A}utoencoder that \textbf{R}etrieves and \textbf{Ge}nerates. We train \marge by self-supervising the \emph{reconstruction} of target text by first \emph{retrieving} a set of related texts (in many languages) and then conditioning on them to maximize the likelihood of generating the original.  We pre-train a multi-source sequence to sequence model that separately encodes each retrieved document and decodes the target, piecing together and translating content from the appropriate inputs as needed to provide the best reconstruction possible. The retrieval model scores are used to bias the cross attention to the most relevant retrieved documents, allowing the retrieval model to be trained jointly from the reconstruction loss.

Our approach can be viewed as a new type of denoising auto-encoder where the noise comes from the retrieval step and is much more diverse than masking; retrieved documents may have little lexical overlap with the target, and may not even be in the same language, but should communicate the same underlying information. 
In this way, the pre-training task is designed to emphasize paraphrasing and reduce the amount of encyclopedic knowledge the model must memorize.
%When this is true, the model can focus on learning to paraphrase, rather than memorizing encyclopedic knowledge.
The set of retrieved documents and relevance scores are an autoencoder bottleneck from which the input must be reconstructed. 
\marge is related to recent work that learns to do retrieval as part of the end task model, for example to find evidence documents in open domain question answering~\citep{realm,rag}. This leads to a more challenging retrieval problem that, unlike ours, requires a separate pre-training phase.

Overall, our pre-trained models capture elements of traditional paraphrasing, translation, multi-document summarization, and information retrieval tasks --- without any fine tuning.\footnote{Masked language models, in contrast, are less directly related to target fine tuning tasks and significant ongoing research focuses on understanding why they work so well, see~\citet{bertology} for a survey.}
%Overall, our pre-training objective captures elements of traditional paraphrasing, translation, multi-document summarization, and information retrieval tasks.
%--- allowing supervised fine-tuning to be a thinner layer on top of pre-training.
 This allows effective zero-shot learning in many cases; with no fine-tuning we achieve BLEU scores of up to 35.8 for document translation, and outperform strong baselines for cross-lingual transfer in summarization. These results provide a step towards pre-trained models that can perform any task with little or no fine-tuning. With fine-tuning, we achieve competitive performance with masked language models on a range of discriminate and generative tasks in many languages, making \marge the most generally applicable pre-training method to date.

%Despite the multi-source training, we also show our model can be effectively fine tuned in a standard sequence to sequence setup with a single encoder, following recent work~\citep{bart,mbart,t5}. We achieve competitive performance with masked language models on a range of discriminate and generative tasks in many languages, making this the most generally applicable pre-training method to date.

\section{Model}
\subsection{Overview}
%We pre-train our model on an unsupervised paraphrasing task. The model learns which documents are most relevant to each other, and how to construct a document from a set of related documents.

During pre-training, the input to the model is a batch of evidence documents\footnote{We use \emph{document} to refer to contiguous chunks of text up to maximum length (here, 512 tokens).} $z_{1..M}$ and target documents $x_{1..N}$. The model is trained to maximize the likelihood of the targets, conditioned on the evidence documents, and the relevance of each evidence document to each target:
\begin{itemize}
    \item The model first computes a relevance score $f(x_i,z_j)$ between every pair of documents $x_i$ and $z_i$, by embedding each document and computing their cosine similarities (\S\ref{section:model_similarity}).
    \item The model then computes the likelihood of reconstructing each $x_i$ conditioned on $z_{1..M}$ and each $f(x_i,\cdot)$, using a modified seq2seq model. The similarity score encourages the model to attend more to relevant evidence documents. Backpropagating the reconstruction loss therefore improves both the sequence-to-sequence model and the relevance model (\S\ref{section:model_reconstruction}).
    \item We construct batches so that evidence documents are relevant to the targets, using the relevance model for retrieval (\S\ref{section:model_retrieval}).
\end{itemize}

Training this model is a chicken-and-egg problem. The reconstruction and relevance models cannot be effectively updated if the batches do not contain relevant evidence documents, but batch construction relies on a relevance model. However, we found that, in practice, the model is able to learn from a random initialization, which effectively provides a type of hashing of random features for each word. 

%From the perspective of an auto-encoder, the encoding of $x_i$ is its relevance scores $f(x_i,\cdot)$.

\subsection{Relevance Scores}
\label{section:model_similarity}
%The relevance model encodes a target document $x_i$ as a set of scalar relevance scores $f(x_i,z_j)$ to a set of evidence documents $z_{1..M}$. 
To learn the relevance scores $f(x_i,z_j)$ for a pair of documents, we train a document encoder $g$ that maps a list of tokens to a fixed size representation. We apply the same encoder to both the target and evidence document, and take the cosine similarity between their representations:
\begin{equation}
    f(x,z)=\frac{g(x)\cdot g(z)}{\norm{g(x)}\norm{g(z)}}
    \label{eq:relevance}
\end{equation}

This function is used in the reconstruction model (\S\ref{section:model_reconstruction}), and trained by the reconstruction loss. It is also used to construct batches of relevant documents (\S\ref{section:model_retrieval}). 

Using the same encoder for both the target and evidence documents allows even random models to compute meaningful similarity functions, as documents with higher lexical overlap are more likely to be projected to more similar representations (\citet{wieting2019} demonstrate this for recurrent models). This property is crucial at initialization.

We encode documents by taking the representation of the first token from the top of a 4-layer Transformer \citep{vaswani:transformer}.
%We implement the encoder by encoding the document with a 4-layer Transformer \citep{vaswani:transformer}, and taking the representation of the first token. 
We share parameters with the first four layers of the reconstruction-model encoder, which saves computation and allows multitask learning.

\subsection{Reconstruction Model}
\label{section:model_reconstruction}

Given a set of evidence documents $z_{1..M}$ and similarity scores $f(x_i,z_j)$, the reconstruction model computes the likelihood of target document $x_i$.

\begin{equation}
    L_\theta=-\sum_i \log p_\theta(x_i|z_{1..M}, f(x_i,z_1),\dots,f(x_i,z_M))
    \label{eq:loss}
\end{equation}

This provides an auto-encoder loss where the reconstruction of document $x_i$ is indirectly conditioned on $x_i$, but with an intermediate bottleneck provided by the retrieved documents and relevance scores, as described in more detail below.

First, the input documents are encoded individually with a bidirectional Transformer, and then the resulting embeddings are concatenated. The similarity score is used to bias the cross-attention from the decoder to the encoder, so that the decoder will pay more attention to more relevant evidence documents. Using more relevant evidence documents will improve the likelihood of reconstructing $x_i$, so gradient descent on (\ref{eq:loss}) will improve the quality of the similarity scores.

%Attention-based sequence-to-sequence models \citep{attention} models compute $a_{uv}=q_uk_v$ for 

Standard Transformer sequence-to-sequence models \citep{vaswani:transformer} compute a matrix of cross-attention probabilities between all elements of target document $x_i$ and evidence document $z_j$:
\begin{equation}
\alpha=\mathit{softmax}_{z_j}(Q^{lh}(x_i)K^{lh}(z_j)) \in \mathbb{R}^{|x_i|\times |z_j|}
\end{equation}
where $Q^{lh}$ and $K^{lh}$ compute query and key representations for layer $l$ and head $h$, and $\mathit{softmax}_{z_j}$ denotes a softmax normalised over elements of $z_j$. 

We instead compute cross attention over a set of evidence documents $z_{1..M}$, biasing the attention scores with the document relevant score from (\ref{eq:relevance}):
\begin{equation}
\alpha=\mathit{softmax}_{z_{1..M}}=Q^{lh}(x_i)K^{lh}(z_{1..M})+\beta f(x_i,z_j) \in \mathbb{R}^{|x_i|\times \sum_j|z_j|}
\label{eq:attention}
\end{equation}

where $\beta$ is a trainable scalar parameter that weights the importance of the document similarity score.

%In the decoder cross-attention layers, we add $\alpha f(x_i,z_j)$ to the standard query-key score for all words in $x_i$ and $z_j$, where $\alpha$ is a trainable scalar parameter. This makes the attention score the sum of word-level and document-level similarities.

\citet{realm} propose a related approach in which the likelihood of a target $x$ is calculated by marginalizing out latent documents $z$: $p(x)=\sum_j p(x|z_j)p(z_j)$. Our attention-like mechanism is (1) more expressive, because it can pay complete attention to a token from one document at one timestep and a token from another document at another timestep, and (2) more efficient because $p(x|z)$ is not computed separately for each $z_j$. However, our method does not allow attention from $z$ to $x$.

%\paragraph{Relationship to REALM} \citet{realm} recently proposed REALM, in which retrieved documents are used to improve a masked language model. In their approach, they optimised the likelihood of a target by marginalising out latent documents $z$: $p(x)=\sum_j p(x|z_j)p(z_j)$. Instead of marginalisation, we use an attention-like mechanism. Our approach has two key advantages: 
%\begin{itemize}
%    \item Firstly, it is more expressive. Different inputs $z$ may be most relevant for different timesteps in the output $x$. The marginalisation approach is normalised over complete documents, so cannot pay complete attention to one document at one timestep and another document at a different timestep.
%    \item Secondly, it is more efficient, as multiple output documents can be decoded in a batch given the same retrieved documents.
%\end{itemize}

%A limitation compared to the REALM approach is that it does not allow cross attention from the retrieved documents to the output.

\subsection{Batch Construction}
\label{section:model_retrieval}
Batches are constructed to create evidence document sets $z_{1..M}$ that give useful information for reconstructing target documents $x_{1..N}$, as detailed in this section. %If the evidence documents contain no relevant information, the model degenerates into an unconditional language model over the targets.
%\paragraph{Overview} 
Overall, we divide the data into \emph{shards} of related documents. Periodically, we compute the similarities between pairs of documents within each shard, using the relevance model, and apply a threshold to keep the strongest connections. The final batches are constructed to maximize connectivity between evidence and target documents.
 
\paragraph{Document similarity}
We compute document similarity in the same way as \S\ref{section:model_similarity}. All documents $x$ are encoded as a vector $g(x)\in \mathbb{R}^d$, and then all pair-wise similarities between documents are computed with a single matrix multiplication.

\paragraph{Data Sharding}
We use simple heuristic constraints to divide documents into related shards, to improve both the accuracy and efficiency of retrieval.
Specifically, for news text, documents are in the same shard iff they were published on the same date.
For Wikipedia, we split articles into chunks of length 512. We create 1000 shards, where all chunks from the same article, or the equivalent article in another language, are in the same shard (otherwise dividing chunks randomly).

% This constraint greatly simplifies the task of finding highly related documents---allowing even a randomly initialized retrieval model to achieve non-trivial accuracy---and also means that the exact nearest neighbor documents can be computed efficiently.

\paragraph{Indexing}
While we backpropagate through the relevance model in (\ref{eq:attention}), the construction of the batch itself is inherently non-differentiable.
For convenience we perform the nearest neighbour search offline. Every 10k model updates, we sample a set of shards of documents. For each shard, we compute $f(x,z)$ for every pair of target and evidence documents, using the current relevance model.  

\paragraph{Thresholding}
We select which documents are sufficiently related by taking the top $k$ most similar document pairs across all pairs in the shard. Some targets may have no sufficiently relevant evidence documents, and are unused until the shard is re-indexed with an updated relevance model. 

\paragraph{Batching}
We aim to construct batches containing clusters of related target and evidence documents, to maximize available information for reconstructing each target.
The output from the thresholding step is a bipartite graph of evidence and target documents with edges between them. A batch is a subgraph, and we perform a small local search to find subgraphs maximizing the sum of the weights of all edges in the subgraph. To encourage the model to build multilingual batches, edges where the evidence and target are in different languages are given weight 100, and other edges have weight 1.
To create batches, we iterate over seed evidence documents $x_i$ with an edge to at least one evidence document. We then greedily add evidence and target documents to the batch to maximize the sum of the weights of edges, until the maximum number of tokens that can fit in GPU memory is reached.

\section{Training}
\paragraph{Architecture}
We use a Transformer model \citep{vaswani:transformer}. The encoder consists of 12 Transformer layers of dimension 1024, with feedforward layers of size 4096.
Recent work showed that large models train more efficiently \citep{trainlarge,scalinglaws}. The decoder is similar to the encoder, but we increase the size of the feed-forward layers in the Transformer decoder to 16536. We also add 4 additional Transformer layers to the base of the decoder with only self-attention and feedforward layers of size 4096, which allows words in the target to contextualize locally before the more expensive cross-attention and feed-forward layers. We focus on scaling up the decoder, because it has access to more information than the encoder (which sees only evidence documents).
In total, the model contains roughly 960M parameters.
For the relevance model, we use the first 4 layers of the encoder, and take the documents representation from the beginning-of-sentence token.
%as a document representation.

\paragraph{Pre-training}
%We use the CC-NEWS corpus \citep{roberta}, which contains news articles published between September 2016 and February 2019, and Wikipedia.

During pre-training, workers process sub-batches containing an average of 2 evidence documents and 2 target documents, and accumulate gradients across workers. Using the CC-NEWS corpus \citep{roberta}, we train initially using the with 64 workers for 450k steps (linearly annealing the learning rate from 1e-04 to 0 with 10k warmup steps), and then continue training with 2048 workers with 550k steps (annealing the learning rate from 2e-04 to 0).\footnote{Initially training with a smaller learning rate reduced instability with an untrained retrieval model.} We refer to this model as \margenews. To explore domain effects, we further pre-train for 100k steps on Wikipedia data, annealing the learning rate from 1e-04 to 0, and refer to the resulting model as \marge.
We rebuild the index every 10k updates. We set retrieval thresholds such that we take on average 4 monolingual and 4 crosslingual links per target document.

\paragraph{Data Pre-processing}
We de-duplicate the data, and identify languages using FastText \citep{fasttext}.
We select documents published in 26 different languages (based on their prevalence in downstream tasks), summarized in the Appendix.
We divide documents into chunks of length 512. We allow all chunks to be evidence documents. For the news domain, we only allow the first chunk in each document to be used as a target, which we found improved performance during development.
We prepend a language identifier token as the first decoder input, to control the output language.

\paragraph{Fine-tuning}
For fine-tuning, we use a similar procedure to \citet{bart}. For generation problems, such as translation and summarization, the task input is fed into the encoder, and the output is generated by the decoder. For classification problems the task input is fed into both the encoder and decoder, and a representation is used from the decoder's final layer hidden state. For zero-shot transfer experiments, we freeze word embeddings and the first 4 decoder layers. %, which improved results.

\section{Experiments}
As a multi-lingual sequence-to-sequence model, \marge is applicable to a very broad range of tasks. We focus on multi-lingual tasks with elements of retrieval, document comprehension, and document generation, because they are the most directly related to our pre-training. %We report \emph{unsupervised zero-shot} performance, in which the pre-trained model is used directly with no fine-tuning, and \emph{cross-lingual transfer}, in which the model is fine-tuned and tested on different languages. 
%Organize by task, or by zero-shot/cross-lingual transfer/fine-tuning?
%\subsection{Comparison Models}
%\paragraph{mBERT}
%\paragraph{mBART}

Table \ref{tab:models} lists the strongest available multilingual pre-trained models, along with relevant model statistics. We compare performance to published numbers for these models. 
%Statistics of comparison models are summarised in T, and include a range of pre-training regimes, representing the strongest available multilingual models.

\begin{table}[]
\scalebox{0.9}{
\begin{tabular}{@{}lrrcrrr}
\toprule
      & \#Parameters         & \#Languages          & Pretraining task     & \begin{tabular}[c]{@{}r@{}}Pretraining GPU\\ Days (estimated)\end{tabular} & \begin{tabular}[c]{@{}r@{}}Pretraining Data\\ (GB; estimated)\end{tabular} \\ \midrule
mBERT  &    172M              &     104     &  MLM                      & Unknown & 60 \\
XLM    &    570M              &     100     &  MLM                      & 640 & 60 \\
XLM-R  &    550M              &     100     &  MLM                      & 27000 & 2394  \\
MMTE   &    192M              &     100     &  Translation              & Unknown & Unknown \\
mBART  &    680M              &      25     &  seq2seq MLM              & 4500   & 1370 \\ 
\midrule
\marge  &    963M              &      26     &  Retrieval+Reconstruction & 4700  & 206  \\ \bottomrule
\end{tabular}
}
\caption{\label{tab:models} Comparison models: \marge is pre-trained on a scale between XLM and XLM-R.}
\end{table}

\begin{table}[]
\vspace{0pt}
\begin{subfigure}{.5\linewidth}
\scalebox{0.75}{
\begin{tabular}{rcccccccc}
\toprule
      & \multicolumn{5}{c}{IWSLT2017} & WMT19                 \\
      & \multicolumn{1}{c}{ar} & \multicolumn{1}{c}{de}& %\multicolumn{1}{l}{es} & 
      \multicolumn{1}{c}{fr}   & \multicolumn{1}{c}{ja} & %\multicolumn{1}{c}{ko} & 
      \multicolumn{1}{c}{zh} & de
      \\ \midrule
%Into English & 25.6 & 27.7 & 37.1 & 4.4 & 16.9   \\
%From English & 6.9 & 18.5  & 32.8 & 0.0 & 0.3  \\       
%\midrule
Into English & 26.8 & 28.5 & 34.3 & 12.6 & 19.9 & 35.8 \\
From English & 12.9 & 14.4 & 25.5 & 10.7 & 12.9 & 13.4 \\       

\bottomrule        

\end{tabular}
}
%\caption{\label{tab:unsupervised_mt} \textbf{Zero-shot unsupervised document level machine translation} BLEU scores using the pre-trained model, with no fine-tuning or special constraints on generation. 
%Performance varies considerably across languages, but is non-trivial with even distantly related languages.
%}
%\end{table}
\end{subfigure}%
\hfill
\begin{subfigure}{.45\linewidth}
\vspace{-10pt}
\scalebox{0.75}{
%\begin{table}[] 
\begin{tabular}{rrcccccc}
\toprule
&& \multicolumn{5}{c}{Target} \\
&       & de  & en   & it   & nl   & ro  \\ 
\midrule
       & de  & -    & 30.6 & 14.0 & 14.8 & 11.6  \\
       & en  & 18.8 & -    & 14.3 & 15.0 & 14.0  \\
Source & it  & 14.0 & 31.7 &  -   & 11.3 & 12.7  \\
       & nl  & 14.3 & 27.5 & 12.6 &    - & 9.3   \\
       & ro  & 14.3 & 32.8 & 14.4 & 9.8  &    -  \\
\bottomrule
\end{tabular}
}
%\caption{\label{table:bucc}\textbf{Zero-shot Unsupervised Document Translation} on parallel text from IWSLT2017. }
\end{subfigure}
\caption{\label{tab:unsupervised_mt} \textbf{Zero-shot unsupervised document level machine translation} BLEU scores using the pre-trained model, with no fine-tuning or special constraints on generation. Performance varies considerably across languages, but is non-trivial with even distantly related languages.
}
\end{table}

%\subsection{Zero-shot and Unsupervised Performance}
%First, we examine the performance of the pre-trained model on downstream tasks, with no task-specific training. These evaluations probe what is being learnt by our pre-training task. We focus on three applications which are most closely related to the pre-training objective: document-level machine translation, document summarization and sentence similarity.

\subsection{Cross-lingual Sentence Retrieval} 
Our pre-training task requires the model to retrieve similar texts, which may be in different languages. 
As an extrinsic evaluation of this functionality, we study cross-lingual sentence retrieval, in which a model must identify the correct translation of a sentence from a set of distractors. We report performance on BUCC2018 \citep{bucc} and Tatoeba \citep{tatoeba}.

We follow the setup of \citet{xtreme}, using no fine-tuning. As a document representation, we use the average embedding of the fifth encoder layer (tuned on BUCC development data).

On BUCC (Table \ref{table:bucc}), \marge outperforms other unsupervised models by almost 10 points. On Tatoeba (see Appendix), there is significant variation across languages, but overall \marge performs comparably to XLM-R and significantly better than other pre-trained models. Better results have been achieved on both tasks using labeled bitext for training \citep{tatoeba}, but our results suggest that our pre-training objective learns an effective cross-lingual retrieval function.

\begin{table}[t] 
\parbox{0.45\textwidth}{
\begin{tabular}{@{}lccccc@{}}
\toprule
      & de   & fr   & ru   & zh   & avg  \\ 
\midrule
mBERT & 62.5 & 62.6 & 51.8 & 50.0 & 56.7 \\
MMTE  & 67.9 & 63.9 & 54.3 & 53.3 & 59.8 \\
XLM   & 56.3 & 63.9 & 60.6 & 46.6 & 56.8 \\
XLM-R & 67.5 & 66.5 & 73.5 & 56.7 & 66.0 \\
\midrule
MARGE & \bf{78.8} & \bf{75.9} & \bf{77.3} & \bf{71.6} & \bf{75.9} \\ 
\bottomrule
\end{tabular}
\caption{\label{table:bucc}Unsupervised Sentence Retrieval results on BUCC. MARGE outperforms other unsupervised models. }
}
\hfill
\parbox{0.5\textwidth}{
\centering
\begin{tabular}{@{}lcc@{}}
\toprule
%& \multicolumn{2}{c}{Source-Target Language Pair} \\
                      & en-de   & zh-en \\ 
\midrule
Random Initialization &   7.7   &   3.2 \\
HAN \citep{han}       &    -    &  24.0 \\
mBART (sentence)      &  38.0   &  28.4 \\
mBART (document)      &  38.5   &  \bf{29.6} \\
\midrule
MARGE                 &  \bf{39.2}  &  28.4 \\
\bottomrule
\end{tabular}
\caption{\label{table:supervised_mt} Supervised document-level machine translation. Comparison results are from \citet{mbart}. \marge performs similarly to mBART.}
}
\end{table}

\subsection{Document-Level Machine Translation} 
During pre-training, the model can retrieve evidence documents in different languages to the target---in contrast to mBERT, XLM and mBART where instances are monolingual. We explore how well this pre-training approach learns to translate. We focus on document level translation tasks, and report document-level BLEU scores.\footnote{All sentences in a document are concatenated prior to calculating BLEU, using SacreBLEU \citep{sacrebleu}.} Following \citet{mbart}, we segment documents into chunks of 512 tokens for training and generation, and then concatenate chunks of the same document.

\paragraph{Zero-Shot Unsupervised Document Translation}
Translation offers a direct measure of how well the pre-trained model encoder and decoder work for different languages, and the extent to which the interface between them is language independent. Therefore, in contrast to prior work on unsupervised translation, we do not further fine-tune the model with iterative back-translation \citep{unsupervisedmt1,unsupervisedmt2}, or bitext in other language pairs \citep{gnmt,mbart}.

We measure both translation into English, which compares encoder performance for other languages, and translation out of English, which measures the decoder performance. Generation hyperparameters were minimally tuned on German/English development, and are shared across all translation pairs. We use a beam of size 6 and block repeated n-grams of length 8 \citep{ngramblocking}.

%Document level translation test sets are only available for a few languages \citep{}. However, many sentence-level datasets were constructed from a concatenation of sentences. We therefore create document-level translation datasets in more languages by contiguous chunks of sentence level test sets.

%\paragraph{baseline} As a baseline, we use mBART \cite{mbart}. mBART is prone to ignoring the language id token, producing outputs in the same language as the source---because, in contrast to \marge, the source and target languages are the same during pre-training. Following Liu et al., we therefore deterministically constrain it to only output tokens that are sufficiently frequent in the target language.

Results are shown in Table \ref{tab:unsupervised_mt}. Performance varies considerably by language, but reaches 35.8 for German to English, which is the highest score we are aware of for system trained with no bitext. Performance is also strong for some languages using different scripts, such as Arabic to English. However, some languages work less well, notably Japanese. Generating non-English languages proves harder in all cases, particularly those with non-Latin alphabets, but English to French works well. Future work should explore up-sampling rarer languages during pre-training. %This may reflect their relative frequency in the training set, which is worthy of more study in future work.

%ome languages, notably \mike{TODO}, work less well. %One possible explanation is that

Qualitatively, we note that the translations are often good but less literal translations than the reference. This may cause BLEU scores to underestimate performance. %Some examples are shown in Table \ref{table:unsupervised_mt_examples}.

It is likely that unsupervised performance could be further improved using iterative back-translation using \marge as an initialization, but we focus here on examining the pre-trained model directly.

\paragraph{Supervised Document Translation}
We also evaluate how well our models can be fine-tuned for translation using labeled bitext. To compare with mBART, we use the same English-German and Chinese-English document translation tasks from WMT19 and IWSLT2015. Table \ref{table:supervised_mt} show that \marge and mBART perform similarly, with \marge performing better on English-German and mBART on Chinese-English. Both outperform baselines by a wide margin.%, suggesting both pre-training methods give strong initializations.

\subsection{Summarization}
We evaluate monolingual sequence-to-sequence generation performance on text summarization tasks.
%We use the standard CNN/DM dataset for English \citep{cnn} and the recent 
We use the MLSum dataset \citep{mlsum} to compare performance in several  languages. % \mike{Are we doing CNN/DM?}
%The source and summary are always in the same language. %, and we treat the problem as a seq2seq generation task.

Results are shown in Table \ref{table:mlsum}. \marge outperforms an extractive mBERT model---the extractive oracle performance suggests that extractive models are very competitive on this dataset---and a seq2seq model without pre-training. 
In some cases, training one model on all languages (train all) improves results.
% We find further gains from training one model on all languages (train all), suggesting that language-independent summarization skills are being shared. 
Finally, we explore zero-shot summarization, where the model is trained on all languages except the test language---this model outperforms a strong lead-3 baseline, and even a supervised pointer-generator model on Spanish and Russian. 
On this domain, we achieve better results with \margenews, a version of the model trained only on news. %To our knowledge, this is the first time a zero-shot summarization model has outperformed a lead-3 baseline.

\begin{table}[]
\centering
\begin{tabular}{@{}rrrrrrrr@{}}
\toprule
                  & \multicolumn{6}{c}{MLSum}             \\
                Model & Setting & de    & es  & fr      & ru    & tr  & avg  \\ \midrule
Extractive Oracle & Oracle & 52.30 & 35.78 & 37.69 & 29.80 & 45.78 & 29.81 \\
Lead 3            & Deterministic & 33.09 & 13.70 & 19.69 & 5.94  & 28.90 & 13.65 \\
Pointer-Generator & Train One & 35.08 & 17.67 & 23.58  & 5.71  & 32.59 & 15.91 \\
M-BERT            & Train One & 42.01 & 20.44 & 25.09 & 9.48  & 32.94 & 17.59 \\
\midrule
%M-BART            &       &       &       &       &       \\
%\marge (zero-shot transfer) & 18.92 & 23.89 & 17.63 & 7.07 &	27.14 \\ 
%\margenews (zero-shot transfer) & 20.68 &	30.56 &	18.36 &	7.97 &	27.80 \\ 
%\marge            & 25.53 &	42.49 &	22.50 &	10.53 &	35.39 \\ 
%\marge (train all) & \textbf{25.88} &	\textbf{43.03} &	\textbf{22.89} &	\textbf{11.07} &	\textbf{35.57} \\

\margenews & Zero-shot Transfer & 30.01 & 17.81 & 19.39 & 8.67 & 29.39 & 15.05 \\
\margenews & Train One & 42.60 & 22.31 & \textbf{25.91} & 10.85 & \textbf{36.09} &  19.03 \\
\marge  & Train All & 42.70 & 22.27 & 25.78 & 10.85 & 35.47 & 18.87 \\
\margenews & Train All & \textbf{42.77} & \textbf{22.72} & 25.79 & \textbf{11.03} & 35.90 & \textbf{19.09}\\

\bottomrule
\end{tabular}
\caption{\label{table:mlsum}
ROUGE-L scores on MLSum. MARGE generates abstractive summaries that outperform an extractive mBERT model. We also demonstrate zero-shot transfer learning, where the model is trained only on languages it is not trained on, and results from training on all languages.
}
\end{table}

\subsection{Paraphrasing}
We measure how well our pre-training task learns paraphrasing on the PAWS-X paraphrase detection dataset \citep{pawsx}. Models must determine whether two sentences are paraphrases; examples were constructed adversarially to have high lexical overlap. Models are trained on English, and we test zero-shot transfer to other languages.  \marge edges out a new state of the art (Table \ref{table:pawsx}).

\subsection{Question Answering}
Question answering offers another document level reasoning task that is easily posed in many languages. We use the MLQA dataset \citep{mlqa}, in which models are trained on the English SQuAD dataset \citep{squad} and then tested in other languages.

Results in Table \ref{table:mlqa} show that MARGE achieves competitive performance with XLM-R, setting the state of the art for Chinese, and outperforms other models by a wide margin.

\begin{table*}[]
\begin{subfigure}{0.5\textwidth}
\scalebox{0.75}{
\begin{tabular}{rcccccccc}
\toprule
& en & ar & de & es & hi & vi & zh & avg \\
  \midrule
mBERT & 80.2 & 52.3 & 59.0 & 67.4 & 50.2 & 61.2 & 59.6 & 61.4  \\
MMTE & 78.5 & 56.1 & 58.4 & 64.9 & 46.2 & 59.4 & 58.3 & 60.3 \\
XLM & 68.6 & 42.5 & 50.8 & 54.7 & 34.4 & 48.3 & 40.5 & 48.5  \\
XLM-R & 83.5 & \bf{66.6} & \bf{70.1} & \bf{74.1} & \bf{70.6} & \bf{74.0} & 62.1 & \bf{71.6}  \\
\midrule
%\marge &     83.4  & 63.5 & 68.3 & 72.7 &67.2 & 70.9 & \bf{67.1}& 70.6 \\
\marge & \bf{83.7} & 64.5 & 68.7 & 73.4 &67.2 & 71.5 & \bf{67.8    }& 71.0     \\
\bottomrule
\end{tabular}
}
\caption{\label{table:mlqa}F1 scores on the MLQA question answering task.
% with XLM-R.%, with a fraction of the pre-training compute.
%Comparison results are taken from \citet{xtreme}.%---with a different fine-tuning procedure \citet{xlmr} reports similar numbers to ours using XLM-R (average 70.7F1).
}
\end{subfigure}%
\hfill
\begin{subfigure}{0.45\textwidth}
\scalebox{0.75}{

\begin{tabular}{cccccccc}
\toprule
en & de & es & fr & ja & ko & zh & avg \\
\midrule
94.0 & 85.7 & 87.4 & 87.0 & 73.0 & 69.6 & 77.0 & 81.9 \\
93.1 & 85.1 & 87.2 & 86.9 & 72.0 & 69.2 & 75.9 & 81.3 \\
94.0 & 85.9 & 88.3 & 87.4 & 69.3 & 64.8 & 76.5 & 80.9 \\
\bf{94.7} & \bf{89.7} & 90.1 & 90.4 & 78.7 &  \bf{79.0} & 82.3 & 86.4 \\
\midrule
\bf{94.7} & 89.4 & \bf{91.6} & \bf{90.9} & \bf{78.9}& 77.7 & \bf{82.5}& \bf{86.5}\\
\bottomrule
\end{tabular}

}
\caption{\label{table:pawsx}Paraphrasing accuracy on PAWS-X.}
\end{subfigure}
\caption{Cross-lingual transfer: models are trained on English (en) and tested on other languages. MARGE performs competitively with XLM-R, with 20\% of the pre-training compute.}
\end{table*}

\section{Analysis}
%\subsection{What does the retrieval model learn?}
% What percentage of links are which language pairs?
% Examples retrievals

\paragraph{What does the reconstruction model learn?}
\begin{table}
%\begin{longtable}{rp{6cm}p{6cm}}
\begin{tabular}{rp{6cm}p{6cm}}
\toprule
Language & Source & Zero-shot Unsupervised Output \\
\midrule
French & Katherine Coleman Goble Johnson, née le 26 août 1918 à White Sulphur Springs (Virginie-Occidentale) et morte le 24 février 2020 à Newport News (Virginie), est une physicienne, mathématicienne et ingénieure spatiale américaine.Elle contribue aux programmes aéronautiques et spatiaux du National Advisory Committee for Aeronautics (NACA) puis de la National Aeronautics and Space Administration (NASA).
& Katherine Coleman Goble Johnson (August 26, 1918 – February 24, 2020) was an American physicist, mathematician, and space engineer. She contributed to the aeronautics and space programs of the National Advisory Committee for Aeronautics (NACA) and later the National Aeronautics and Space Administration (NASA). \\
Spanish & 
Katherine Coleman Goble Johnson (White Sulphur Springs, Virginia Occidental; 26 de agosto de 1918 - Newport News, Virginia; 24 de febrero de 2020) fue una física, científica espacial y matemática estadounidense que contribuyó a la aeronáutica de los Estados Unidos y sus programas espaciales con la aplicación temprana de las computadoras electrónicas digitales en la NASA. Conocida por su precisión en la navegación astronómica, calculó la trayectoria para el Proyecto Mercury y el vuelo del Apolo 11 a la Luna en 1969.
& 
Katherine Coleman Goble Johnson (August 26, 1918 – February 24, 2020) was an American physicist, space scientist, and mathematician who contributed to the United States aeronautics and its space programs with the early application of digital electronic computers at NASA. Known for her accuracy in astronomical navigation, she calculated the trajectory for Project Mercury and the Apollo 11 flight to the Moon in 1969.
\\
German & 
Katherine G. Johnson (gebürtig Coleman, zwischenzeitlich verheiratet Goble; * 26. August 1918 in White Sulphur Springs, West Virginia; † 24. Februar 2020 in Newport News, Virginia) war eine US-amerikanische Mathematikerin afroamerikanischer Abstammung. Für ihre Beiträge zur Berechnung der Flugbahnen für das Mercury-Programm und den ersten bemannten Flug zum Mond im Rahmen der Apollo-11-Mission wurde sie Ende 2015 mit der Presidential Medal of Freedom ausgezeichnet.
& 
Katherine G. Johnson (née Coleman; August 26, 1918 – February 24, 2020) was an American mathematician. She was awarded the Presidential Medal of Freedom in 2015 for her contributions to the calculation of the flight paths for the Mercury program and the first manned flight to the Moon in the Apollo 11 mission.\\

Italian & 
Katherine Coleman Goble Johnson (White Sulphur Springs, 26 agosto 1918 – Hampton, 24 febbraio 2020) è stata una matematica, informatica e fisica statunitense. Contribuì alla scienza dell'aeronautica statunitense e ai programmi spaziali, già dal primo utilizzo dei computer elettronici digitali da parte della NASA. Venne molto apprezzata per l'accuratezza che poneva nel calcolo della navigazione spaziale computerizzata e per il lavoro tecnico dirigenziale pluridecennale svolto alla NASA: da quando calcolava le traiettorie delle orbite, paraboliche e iperboliche, le finestre di lancio e i percorsi di ritorno di emergenza per molti voli, al Project Mercury, incluse le prime missioni NASA di John Glenn, Alan Shepard, le traiettorie di inserzione lunare nei voli lunari del programma Apollo, continuando con il lavoro sul programma dello Space Shuttle, infine con la progettazione dei primi piani per la missione su Marte.
&
Katherine Coleman Goble Johnson (White Sulphur Springs, August 26, 1918 – Hampton, February 24, 2020) was an American mathematician, computer scientist, and physicist. She contributed to the science of the U.S. Air Force and space programs, as early as the first use of digital electronic computers by NASA. She was highly regarded for the accuracy she put into computerized space navigation calculations and for the decades-long technical leadership work she performed at NASA: from calculating orbital trajectories, parabolic and hyperbolic, launch windows, and emergency return paths for many flights, to Project Mercury, including the first NASA missions of John Glenn, Alan Shepard, lunar insertion trajectories in the Apollo lunar flights, continuing work on the Space Shuttle program, and finally designing the initial plans for the Mars mission.
\\
\bottomrule
\end{tabular}
\caption{\label{table:unsupervised_mt_examples} Example zero-shot unsupervised inputs and outputs (truncated for clarity).}
\end{table}
%\end{table}

To build intuitions about what the reconstruction model learns, we examine model outputs for inputs in different languages on the same topic (Table \ref{table:unsupervised_mt_examples}).

Even for a fixed topic, the model output varies significantly with the input, showing that it is not simply memorizing text. 
Almost all facts in the outputs are supported by the input, with few hallucinations---suggesting pre-training has taught the model to translate and paraphrase information from its source, rather than memorize facts in its parameters.
However, the outputs are not literal translations of the input---in particular, some important facts from the source are not expressed in the output. 
%However, the outputs are not literal translations of the input---some facts are missing; some sentences are re-ordered or merged. Overwhelmingly, however, facts in the output are supported by the input, suggesting that the model has learned to translate and paraphrase information from its source.

The model was not trained on literal translations, so it is perhaps surprising that the output is often so closely aligned to the input. One possible explanation is that more literal translations represent a mode of a diverse distribution over paraphrases.

\begin{figure*}[t!]
    \centering
        \vspace{-2cm}
    \includegraphics[width=\textwidth]{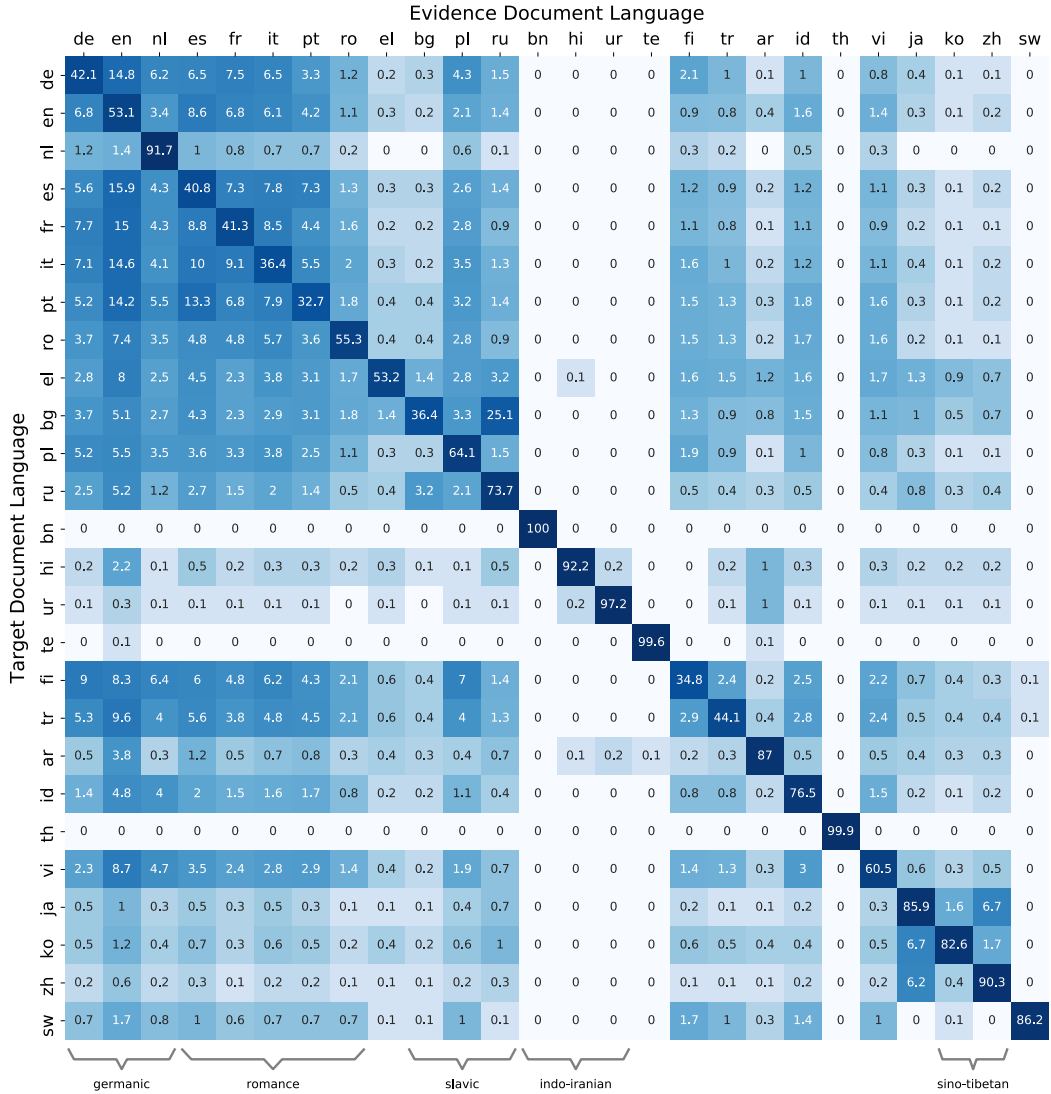}
    \vspace{-2cm}
    \caption{Percentage of retrieved links to documents in target languages (y-axis) from evidence documents in different source languages (x-axis) on Wikipedia. 
    }
    \label{fig:retreival_by_language}
\end{figure*}

\paragraph{What does the retrieval model learn?}
Figure \ref{fig:retreival_by_language} shows statistics of the retrieval model. Differences across languages are due to many factors, including the frequency of  languages in the corpus, how linguistically related two languages are, and how likely two languages are to cover the same topic. Our pre-training also introduces feedback loops, because if the reconstruction model is unable to translate between two languages, it may train the retrieval model that documents in these languages are less relevant to each other.

All languages retrieve the highest proportion of documents within their own language (represented by the diagonal), but otherwise the retrieved documents tend to be distributed over a number of other languages. There tend to be closer affinities between geographically or linguistically related languages, such as Bulgarian and Russian, or Chinese and Japanese. For some languages, the model fails to retrieve many documents in other languages---particularly Indo-Iranian languages, and those which are the only example of their language family we include (such as Telugu and Thai). 
%For these cases, the pre-training effect will be closer to prior multilingual models such as mBART, mBERT, and XLM.
For these cases, the pre-training reduces to independent updates for each language, as used in multilingual models such as mBART, mBERT, and XLM.

% Example translation
% Learning curves by language?

\paragraph{Discussion}
Overall, \marge shows strong performance on a wider range of tasks than any previous pre-trained models, and is effective at discriminative and generative tasks in many languages. Results are competitive with less general models, even XLM-R, which was trained with significantly higher pre-training resources. The pre-training task is more closely related to downstream tasks than masked language modeling, allowing pre-trained models to achieve BLEU scores as high as 35.8 for translation. \marge also broadens the range of known effective pre-training tasks beyond MLMs, which we hope will lead to further exploration and understanding of pre-training objectives.

However, there are several limitations that future work should address. We pre-trained on news and Wikipedia, where simple metadata can be used to constrain the similarity search, improving efficiency and accuracy. Broadening the domains may require approximate nearest neighbor search \citep{faiss}. Learning the retrieval model requires batch sizes greater than one, so model-parallel training would be required to train significantly larger models. Finally, performance is inconsistent across languages, which may be due to feedback loops during training where documents in less well performing languages may learnt to be less relevant, and therefore retrieved less often.

\section{Related Work}

\paragraph{NLP pre-training} Since BERT \citep{bert}, pre-training for NLP has been dominated by variants of masked language models. For example, \citet{xlnet} predicts the masked tokens auto-regressively, \citet{unilm} multitasks MLM and language modeling objectives, \citet{electra} trains a discriminator to classify the correctness of MLM samples, and \citet{bart} and  \citet{t5} use seq2seq models with masked inputs. \marge departs significantly from these objectives in that the inputs during pre-training are complete, uncorrupted text. % We hope that the effectiveness of \marge both motivates exploration of a wider range of pre-training objectives, and leads to better understanding of what methods work and why.

\paragraph{Bitext Mining} Recent work has shown impressive results on machine translation through bitext mining \citep{ccmatrix}, in which a retrieval model is used to search for parallel sentences in a large multilingual corpus, which are then used as training data for a machine translation model. A key conceptual difference is that literal bitext is not optimal for our approach, as we hope to learn linguistic information by training on noisy document-level paraphrases. We also learn to retrieve and translate with no manually translated sentences, unlike existing bitext mining methods.
%We therefore work with documents, whereas bitext mining works with sentences. Bitext mining work uses manually translated sentences to learn the retrieval model, whereas our model learns to retrieve and translate with no manually translated sentences.

\paragraph{Cross-lingual Learning} Several attempts have been made to pre-train language-independent representations. One strand uses MLMs on the concatenation of monolingual corpora, relying on parameter sharing to learn cross-lingual representations \citep{xlm,xlmr,mbart}. Another strand has trained machine translation systems \citep{cove,mmte}, but results in \citet{xtreme} suggest translation is a less effective pre-training task. We instead pre-train on loose cross-lingual paraphrases.

\paragraph{Language Models with Retrieval} 
Several recent papers have shown that word prediction can be improved by retrieving relevant evidence documents.
\citet{realm} and \citet{rag} improve MLMs and text generation by learning to retrieve relevant evidence documents.
\citet{guu2018} perform language modeling by retrieving and editing sentences.
%Most relevantly, the prototype editing framework of \citep{guu2018} also uses retrieved documents to improve a language model. Where we jointly learn a retrieval model, their approach maximizes the unconditional likelihood of the target by marginalises over evidence documents.
kNN-LM \citep{knnlm} shows that language models can be improved with retrieving from the training set, by interpolating a language model with a nearest neighbor classifier. 
In contrast, we learn retrieval during training but do not require it for inference. 
Perhaps most relevantly, \citet{generatingwikipedia} generate Wikipedia articles conditioned on a set of evidence documents. 
%Our work differs in retrieving learning the retrieval model,
%\mike{Generating Wikipedia by Summarizing...}

\section{Conclusion}
We introduced a new approach to pre-training models for natural language understanding and generation, by using retrieved documents to reconstruct the original. \marge exhibits strong performance on a range of discriminative and generative tasks in many languages, both with and without fine-tuning. These results establish \marge as a viable alternative to masked language modeling and provide a step towards pre-trained models that can perform any task with little or no fine-tuning. Future work should scale \marge to more domains and languages, and study how to more closely align pre-training objectives with different end tasks.

%\begin{ack}
%\end{ack}
%\theendnotes
%\printendnotes

\newpage
\bibliographystyle{plainnat}
\bibliography{bibliography}

\newpage

\appendix
\section{Additional Results}

\begin{table}[h]
\begin{tabular}{rcccccccccccc}
\toprule
& ar & bg & bn & de & el & es & fi & fr & hi & id & it & ja \\
\midrule
%BERT & 62.5 & 62.6 & 51.8 & 50.0 & 56.7 \\
%XLM  & 56.3 & 63.9 & 60.6 & 46.6 & 56.8 \\
XLM-R  & 47.5 & \bf{71.6} & \bf{43.0} & 88.8 & \bf{61.8} & 75.7 & \bf{71.6} & 73.7 &  \bf{72.2} &  \bf{77.0} & 68.3 & \bf{60.6} \\
%MMTE & 67.9 & 63.9 & 54.3 & 53.3 & 59.8 \\
\midrule
\marge & \bf{49.9} & 70.5 & 16.9 & \bf{88.9} & 57.2 & \bf{82.9} & 55.8 & \bf{77.0} & 67.1 & 73.8 & \bf{76.5} & 60.1  \\
\\
\toprule
& ko & nl & pt & ru & sw & te & th & tr & ur & vi & zh & \\
\midrule
XLM-R & \bf{61.4} & 80.8 & 82.2 & 74.1 & 20.3 & \bf{35.9} & 29.4 &  \bf{65.7} & 24.3 & 74.7 & 68.3 \\
\midrule
\marge & 50.6 & \bf{84.3} & \bf{84.8} & \bf{78.7} & \bf{22.8} & 16.2 & \bf{38.0} & 63.2 & \bf{41.9} & \bf{77.3} & \bf{77.2} \\
\bottomrule
\end{tabular}
\caption{\label{table:tatoeba}\textbf{Tatoeba} zero-shot sentence retrieval results. \marge performs comparably to XLM-R, but with significant variation across languages. We only show results for languages in all model's pre-training data.}
\end{table}

\section{Pre-training Data}
\begin{table}[h]
\centering
\begin{tabular}{rrrrr}
\toprule
Language & Code & Language Family & CCNews & Wikipedia \\
\midrule
Arabic & ar   & Afro-Asiatic &  2416996 & 747891 \\
Bulgarian & bg  & Slavic &  496023 & 297989 \\
Bengali & bn  & Indo-Iranian & 741  & 134560 \\
German & de  & Germanic& 13320055  & 2735591 \\
Greek & el  & Hellenic & 1793198  & 317780 \\
English & en  & Germanic & 57061325   & 6372976 \\
Spanish & es  & Romance &  16990991 & 2111406 \\
Finnish & fi  & Uralic &   471029 & 496988 \\
French & fr  & Romance & 7281926  &  2749382 \\
Hindi & hi  & Indo-Iranian &  1907850  & 124816 \\
Indonesian & id  & Austronesian & 1295060 & 435599 \\
Italian & it  & Romance &  6865752 & 1776998 \\
Japanese & ja  & Japonic & 458675  & 1311915 \\
Korean & ko & Sino-Tibetan &  1241560 & 442675 \\
Dutch & nl  & Germanic & 2091796  & 1359535 \\
Polish & pl  &  Slavic & 1153817 & 1219494 \\
Portuguese & pt  & Romance & 2971009  & 1107798 \\
Romanian & ro  & Romance & 1960236  &  348036\\
Russian & ru  & Slavic & 6579113 & 1939546 \\
Swahili & sw  & Niger-Congo & 11878  &  34107\\
Telugu & te  & Dravidian &  7155  & 80131 \\
Thai & th  &  Kra-Dai & 5412  &  156505\\
Turkish & tr  & Turkic & 3524089  & 353028 \\
Urdu & ur  & Indo-Iranian&  154912 & 96773 \\
Vietnamese & vi  & Austro-Asiatic & 1019445  & 566375 \\
Chinese & zh  & Sino-Tibetan & 434378  & 1027950 \\
\bottomrule
\end{tabular}
\caption{\label{table:data} Number of documents per language used for pre-training. Languages represent a range of families and geographical regions. The Germanic, Hellenic, Romance, Slavic, and Indo-Iranian families are part of a broader Indo-European family. }
\end{table}

%\section{Example Pre-training Batches}

%\begin{table}[]
%\centering

%\section{Example Zero-shot Unsupervised Translations}

%\newpage

%\section*{Broader Impact}
%This work has a broad scope, covering discriminative and generative tasks in many languages. As such, the broader impact is similar to that of the field of NLP; there exist many potential good and bad applications. Our work learns to translate languages in an unsupervised way, which could be used to bring NLP to more languages but could also potentially introduce more translation mistakes that supervised methods.
%Authors are required to include a statement of the broader impact of their work, including its ethical aspects and future societal consequences. Authors should discuss both positive and negative outcomes, if any.
\end{document}